\documentclass[runningheads]{llncs}

\usepackage{eccv}

\usepackage{eccvabbrv}

\usepackage{graphicx}
\usepackage{booktabs}
\usepackage{wrapfig}

\usepackage[accsupp]{axessibility}  

\usepackage{hyperref}

\usepackage{orcidlink}

\usepackage{soul}
\usepackage{booktabs}
\usepackage{verbatim}
\usepackage{graphicx}
\usepackage{tikz}
\usetikzlibrary{arrows.meta, positioning, fit, calc}

\newcommand{\figfont}{\scriptsize}
\newcommand{\figtiny}{\tiny}

\definecolor{beeorange}{RGB}{230,110,25}
\definecolor{pollenyellow}{RGB}{232,174,35}
\definecolor{pollenbrown}{RGB}{130,78,42}
\definecolor{hiveyellow}{RGB}{232,174,35}
\definecolor{hivebrown}{RGB}{130,78,42}
\definecolor{chamberblue}{RGB}{55,105,165}
\definecolor{stageblue}{RGB}{50,90,145}
\definecolor{stagegreen}{RGB}{70,140,80}
\definecolor{stagepurple}{RGB}{115,85,150}
\definecolor{lightgraybg}{RGB}{248,248,248}

\newcommand{\BeeIcon}[1][0.25]{%
\begin{tikzpicture}[scale=#1, baseline=-0.5ex,
  wing/.style={fill=blue!12, draw=black!25, line width=0.25pt},
  fuzz/.style={draw=black!45, line width=0.18pt, line cap=round}
]

  \path[wing, rotate around={22:(-0.18,0.22)}]
    (-0.18,0.22) ellipse (0.34 and 0.16);
  \path[wing, rotate around={-22:(0.18,0.22)}]
    (0.18,0.22) ellipse (0.34 and 0.16);

  \fill[black!82] (0,-0.38) ellipse (0.34 and 0.50);

  \fill[pollenyellow!95!orange] 
    (-0.30,-0.08)
    .. controls (-0.18,0.05) and (0.18,0.05) .. (0.30,-0.08)
    .. controls (0.25,-0.18) and (-0.25,-0.18) .. (-0.30,-0.08)
    -- cycle;

  \draw[black!25, line width=0.25pt] (-0.26,-0.25) .. controls (0,-0.17) .. (0.26,-0.25);
  \draw[black!25, line width=0.25pt] (-0.23,-0.43) .. controls (0,-0.35) .. (0.23,-0.43);
  \draw[black!25, line width=0.25pt] (-0.18,-0.59) .. controls (0,-0.53) .. (0.18,-0.59);

  \fill[pollenyellow!95!orange] (0,0.18) ellipse (0.38 and 0.34);

  \fill[black!75]
    (0,0.34)
    .. controls (-0.08,0.25) and (-0.07,0.11) .. (0,0.05)
    .. controls (0.07,0.11) and (0.08,0.25) .. (0,0.34)
    -- cycle;

  \fill[black!85] (0,0.62) ellipse (0.24 and 0.22);

  \fill[pollenyellow!85!orange] (0,0.76) ellipse (0.10 and 0.04);

  \draw[black!70, line width=0.25pt]
    (-0.08,0.80) .. controls (-0.22,0.98) .. (-0.34,0.96);
  \draw[black!70, line width=0.25pt]
    (0.08,0.80) .. controls (0.22,0.98) .. (0.34,0.96);

  \foreach \x/\y/\dx in {-0.24/0.10/-0.38, -0.25/-0.12/-0.40, -0.20/-0.35/-0.34,
                          0.24/0.10/0.38, 0.25/-0.12/0.40, 0.20/-0.35/0.34} {
    \draw[black!70, line width=0.25pt]
      (\x,\y) .. controls (\dx,\y-0.08) .. (\dx,\y-0.22);
  }

  \foreach \a in {200,220,240,300,320,340} {
    \draw[fuzz] ({0.34*cos(\a)},{0.18+0.31*sin(\a)})
      -- ({0.42*cos(\a)},{0.18+0.38*sin(\a)});
  }
  \foreach \a in {210,230,250,270,290,310,330} {
    \draw[fuzz] ({0.30*cos(\a)},{-0.38+0.47*sin(\a)})
      -- ({0.37*cos(\a)},{-0.38+0.55*sin(\a)});
  }

\end{tikzpicture}%
}

\newcommand{\ChamberIcon}[1][0.25]{%
\begin{tikzpicture}[scale=#1, baseline=-0.5ex]

  \draw[
    fill=chamberblue!18,
    draw=chamberblue,
    line width=0.7pt,
    rounded corners=0.12cm
  ]
    (-0.42,-0.32) rectangle (0.42,0.32);

  \draw[
    fill=chamberblue!55,
    draw=chamberblue!70!black,
    line width=0.45pt,
    rounded corners=0.08cm
  ]
    (-0.33,-0.25) rectangle (0.33,0.25);

  \draw[
    draw=white,
    line width=0.25pt,
    rounded corners=0.05cm,
    opacity=0.75
  ]
    (-0.24,-0.17) rectangle (0.24,0.17);

\end{tikzpicture}%
}

\newcommand{\HiveIcon}[1][0.25]{%
\begin{tikzpicture}[scale=#1, baseline=-0.5ex]
  \foreach \x/\y in {0/0,0.36/0,-0.36/0,0.18/0.31,-0.18/0.31,0.18/-0.31,-0.18/-0.31}{
    \draw[fill=hiveyellow!85, draw=hiveyellow!40!black, line width=0.25pt]
      (\x+0.18,\y) -- (\x+0.09,\y+0.16) -- (\x-0.09,\y+0.16) --
      (\x-0.18,\y) -- (\x-0.09,\y-0.16) -- (\x+0.09,\y-0.16) -- cycle;
  }
\end{tikzpicture}%
}

\newcommand{\PollenIcon}[1][0.25]{%
\begin{tikzpicture}[scale=#1, baseline=-0.5ex]
  \fill[pollenbrown!25] (0,0) circle (0.35);
  \draw[pollenyellow, line width=0.7pt] (0,0) circle (0.35);
  \fill[pollenyellow!60] (0,0) circle (0.23);
\end{tikzpicture}%
}

\usepackage{pgfplots}
\usepackage{pgfplotstable}
\pgfplotsset{compat=1.18}
\usepackage{svg}

\usepackage{multirow}

\newif\ifshowfigures
\usepackage{environ}

\NewEnviron{maybefigure}[1][]{%
  \ifshowfigures
    \begin{figure}[#1]
      \BODY
    \end{figure}
  \fi
}

\showfigurestrue

\begin{document}

\title{BeeWhere: Segmenting Bumble Bee Colonies to Quantify Behavioral Effects}

\titlerunning{BeeWhere}

\author{Roberta Hunt\textsuperscript{*}\inst{1,2}\orcidlink{0000-0003-1963-3281} \and
August Easton-Calabria\textsuperscript{*}\inst{3}\orcidlink{0000-0002-3373-3302} \and
James Crall\inst{3}\orcidlink{0000-0002-8981-3782}}
\authorrunning{R.Hunt and A.Easton-Calabria et al.}

\institute{University of Copenhagen, 2100 Copenhagen, Denmark \and
Lund University, 223 62 Lund, Sweden \and
University of Wisconsin-Madison, Madison, WI 53706, USA}

\maketitle
\begingroup
\renewcommand{\thefootnote}{\fnsymbol{footnote}}
\footnotetext[1]{These authors contributed equally to this work.}
\endgroup

\begin{abstract}
    Social bees are important pollinators that support biodiversity and crop pollination globally and serve as important model systems for collective behavior, but scalable measurement of individual- and colony-level behavior remains difficult in dense, occluded nest environments. Existing monitoring workflows use fiducial tags (e.g., ArUco) to preserve individual identity, yet tag-based tracking can fail when markers are obscured and provide limited information about body extent, spatial context, and untagged individuals. We present BeeWhere, an AI-assisted annotation and analysis workflow that combines ArUco detections with deep-learnt instance segmentations to quantify bumble bee behavior from high-resolution colony images and videos. Using bumble bee (\textit{Bombus impatiens}) microcolonies as a test case, we annotate 483 frames containing 8,443 bee instances. We additionally annotate pollen balls, nest structures, and chamber boundaries, and train YOLO instance segmentation models for downstream behavioral analysis. Instance segmentations enable quantification of important behavioral metrics based on body contours, including nearest-neighbor distance, proximity to nest structures, spatial occupancy within the nest, and detection counts (proxy for colony growth) over time. We apply the BeeWhere models to tag-based (ArUco) tracking in an exploratory validation study assessing the behavioral impacts of neonicotinoid pesticide exposure. BeeWhere increased detection rates compared to tag-based tracking, particularly when bees were partially obscured (i.e. on the nest structure) or under challenging (blurry) imaging conditions, and also captured treatment-associated changes in bee spatial organization (with pesticide-exposed microcolonies showing greater distance between bees consistent with previous findings) not captured using tag-based tracking alone. These results suggest that instance segmentation can complement fiducial-marker tracking by recovering behaviorally meaningful signals under challenging colony conditions. 
\end{abstract}

\section{Introduction}
\label{sec:intro}
\begin{maybefigure}[tb]
  \centering
  \includegraphics[width=0.9\textwidth]{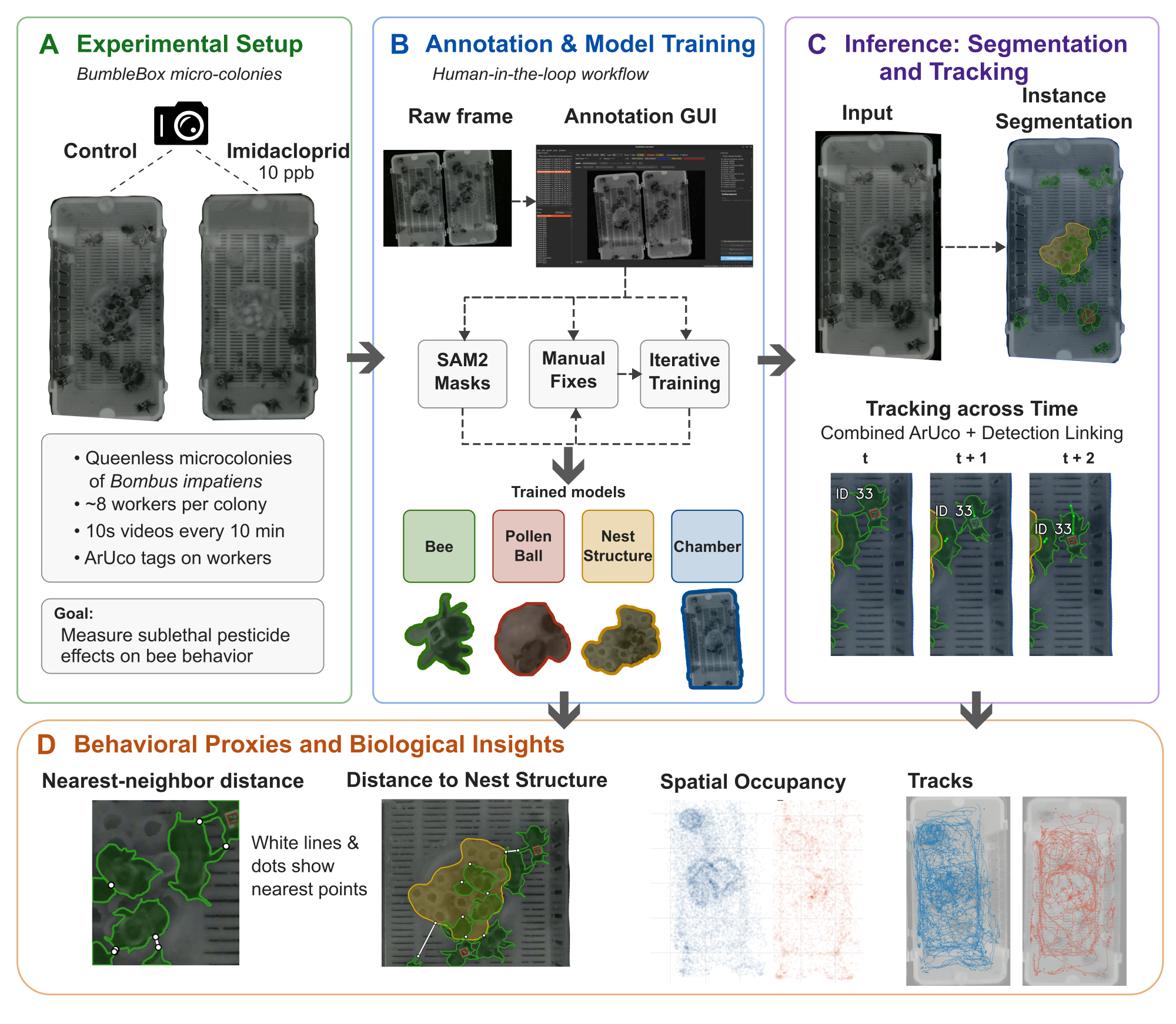}
  \caption{Overview of BeeWhere pipeline. (a) Bumblebox microcolonies of \textit{Bombus impatiens} are imaged under control and imidacloprid treatment conditions. (b) An AI-assisted annotation workflow produces training data for bees, nest structures, pollen balls, and chambers using SAM2 mask proposals and iterative YOLO retraining. (c) Trained models generate instance segmentations that can be linked across time with tracking and ArUco identity information. (d) Segmentation and tracking outputs are converted into behavioral proxies to compare treatments.}
  \label{fig:beewhere}
\end{maybefigure}
Animal pollinators are foundational to biodiversity and agricultural production, contributing to the reproduction of wild flowering plants and sustaining crop pollination services that support food security, livelihoods, and ecosystem stability \cite{Potts2010,Potts2016,Dicks2021}. Social bees (including bumble bees and honey bees) are particularly important pollinators, and are also a key model system in behavior and ecology. Scalable approaches for tracking bees and quantifying behavior can therefore provide practical tools for pollinator health and conservation, while also enabling fundamental research on social behavior and collective behavior.

Social bees and other pollinators are increasingly exposed to multiple, interacting global change stressors, including land-use change, climate change, pathogens, and pesticides, with mounting evidence that these pressures contribute to declining pollinator populations \cite{janousek2023recent} and pollination services \cite{stanley2015neonicotinoid}. A major challenge, however, is that many of these stressors have complex effects on behavior or physiology that may occur well below acutely lethal levels of stress.  Pesticides, for example, can alter behavior, physiology, and reproduction and have long-term colony- and population-level consequences at doses well below lethal toxicity \cite{Raine2024,Crall2018,Pamminger2025}. However, detecting such 'sublethal' effects remains challenging and behavioral assays have historically relied on manual observations that limit scalability and standardization across studies. Although recent advances in computer vision and high-throughput “ethomics” have begun to expand the scope of behavioral phenotyping in bees,  \cite{Smith2022,Jain2025}, quantifying behavior in cluttered, dynamic, and socially complex settings typical of naturalistic applications remains a substantial bottleneck. Closing this methodological gap is important not only for understanding the basic ecology and behavior of social bees, but also to support biodiversity conservation and ecosystem services by providing standardized approaches that can directly inform environmental risk assessment for pesticides and other environmental stressors. \cite{Fisher2023}. .

Here, we develop a hybrid deep-learning segmentation and tracking approach for detecting and tracking bumble bees (\textit{Bombus impatiens}) within nest chambers. Our approach expands on existing tracking approaches within social bee colonies to provide markerless detection of workers \textit{via} instance segmentation to quantify behavior and colony growth. It also provides integration with fiducial markers (ArUco) for long-term individual identity retention, and segmentation of structural nest elements (e.g. developing brood) within the experimental chambers to characterize key aspects of behavior such as nursing. To achieve this, we built a human-in-the-loop annotation tool for the data collected using the BumbleBox system, and applied initial annotations from that to single frames and videos from experiments run over 72 hours in a validation experiment assessing the effects of pesticide exposure on behavior. From this we can study in more detail the behavioral effects of pesticides within social bee colonies.  

\subsubsection{Contributions} The contributions of this paper are fourfold, we provide: 1) a training and validation dataset of 483 object detection frames and 13 tracking sequences across 122 frames annotated with instance segmentations and/or bounding boxes on data using the standardized BumbleBox setup \cite{ref:easton2025bumblebox} and previous microcolony setups, 2) trained YOLO models which can be used on a variety of similar microcolony setups, 3) a GUI for further annotating the videos and retraining models, and 4) we demonstrate the utility of this tool by applying it to a dataset of 3,761 videos across 3 rounds of imidacloprid dosing, showing that instance segmentations combined with ArUco can improve tracking and behavioral metrics. Source code is available at \url{https://github.com/robertahunt/bumblebox_annotator}, trained models, and annotations are available on zenodo (DOI: 10.5281/zenodo.21870158).

\section{Background}
\label{sec:background}

\subsection{Current Bee Tracking Methodologies and their Limitations}
Automated tracking of bees has advanced substantially in recent years, but the most reliable systems for tracking in dense nest environments still tend to preserve identity using externally visible, fiducial tags in both bumble bees \cite{crall2015beetag} and honey bees \cite{Wario2015, Gernat2018PNAS}. In bumble bees (\textit{Bombus} spp), the BumbleBox system uses ArUco markers for individual tracking in a custom nest imaging chamber \cite{ref:easton2025bumblebox} and has been used to monitor how pesticide exposure alters critical behaviors (e.g. thermoregulatory clustering and nursing) \cite{EastonCalabria2023}.  ArUco markers \cite{garrido2014automatic} are standardized fiducial tags, which give information about the 3 dimensional orientation of the ArUco tag relative to the camera. They are widely used in robotics\cite{kalaitzakis2021fiducial}. Fiducial tags have seen recent adoption to animal monitoring and have been successfully used to monitor bees, birds and other animals \cite{VANPUTTEN2025100703,sadrzadeh2024automated,othayoth2022terrain,crall2015beetag}. 

While this approach has been effective and provides reliable re-identification, tag-based tracking alone has important limitations. Tags are often obscured (e.g. especially within large, high-density colonies), are lost or dirtied over time, and do not give any information about the pose or extent of the bee and their surroundings. Tag-based tracking systems also require (often intensive) manipulation of workers and colonies to apply tags, and are not able to track any bees that eclose during the observation time. Given these limitations, markerless tracking with deep learning \cite{Mathis2018DeepLabCut, Pereira2022SLEAP, RomeroFerrero2019idtracker} is a promising approach that is increasingly adopted for bee tracking \cite{Bozek2018CVPR, Bozek2021NatComm}. However, markerless tracking approaches also have important limitations for tracking bees, including retention of identity, especially over extended time periods or when individuals come in and out of the observation area. Recent approaches have integrated markerless, pose-based tracking with fiducial tags to retain individual identity over extended time periods \cite{Smith2022, wolf2023naps}, but these deep-learning pose-based approaches still have limited performance under highly dense conditions within hives or when animals are partially obscured by three dimensional structures (such as brood or other nest materials) as is common in natural colony conditions.

\subsection{Effects of Pesticides on Bumble Bees}
Neonicotinoid insecticides can alter worker behavior within bumble bee nests and contribute to declines in colony growth and reproduction. Automated tracking of individually identified workers with fiducial tags has shown that field-realistic exposure to a common neonicotinoid pesticide (imidacloprid) reduces worker activity and nursing and alters nest occupancy and spatial dynamics (with workers shifting to the nest periphery, away from developing brood)  \cite{Crall2018}. Neonicotinoid exposure can disrupt brood development, suggesting that altered behavior of workers and queens can propagate to colony growth \cite{Chole2022}, with well-documented effects of neonicotinoid exposure on whole colony growth \cite{Whitehorn2012, FauserMisslin2014}.

While there is thus well-established evidence for adverse effects of sublethal pesticide exposure on bee behavior, significant technical and knowledge gaps still exist. The majority of empirical studies focus on one or a small number of active pesticide ingredients at a single dose or concentration. However, bees are exposed to an enormous diversity of agrochemicals in natural and agricultural environments at exposure levels that can fluctuate substantially seasonally and with landscape context \cite{traynor2016hive, nicholson2024pesticide}. Interactions between agrochemicals \cite{siviter2021agrochemicals}, as well as with secondary stressors (such as temperature \cite{EastonCalabria2023}, pathogens \cite{di2013neonicotinoid}, or nutrition \cite{tosi2017neonicotinoid}), can also significantly modulate the direct effects of exposure. These complexities mean that the  impacts depend strongly on dose, environmental context, and cocktail of agrochemicals present during exposure and contribute to the challenge of predicting pesticides' effects in natural landscapes \cite{woodcock2017country}. Scalable approaches such as automated behavioral phenotyping may help address this key challenge.

\subsection{Object Detection and Instance Segmentation}
Object detection and instance segmentation are central tasks in computer vision that have advanced rapidly over the past decade, driven in part by the development of efficient architectures such as YOLO (You Only Look Once)\cite{redmon2016you}. Object detection aims to locate and classify objects within images, typically by predicting bounding boxes and class labels. A survey of recent object detection methods can be found in \cite{arkin2023survey}. Importantly, object detection distinguishes between separate instances of an object class by assigning each instance an individual detection.

Instance segmentation extends object detection by providing a pixel-level mask for each detected object instance. Rather than representing objects only by bounding boxes, instance segmentation estimates the spatial extent of each object, allowing more detailed downstream analyses of object shape, position, and interactions. An example of this distinction is shown in \cref{fig:beewhere}, where individual instances have their own outlines, even when directly connected.

\subsection{Object Tracking in Videos}
Object tracking in video aims to keep consistent identities by linking detections across frames. Tracking methods range from simple centroid-based matching, where detections are linked based on how close they are in space, to overlap-based methods such as intersection-over-union (IOU) matching \cite{bochinski2017high}. Newer multi-object tracking methods, including ByteTrack \cite{zhang2022bytetrack} and BoT-SORT \cite{aharon2022bot}, improve tracking by using both high- and low-confidence detections, which can reduce missed links and broken tracks. A broader summary of recent tracking methods are provided by Li et al.\cite{li2025review}. However, many modern methods still rely at least partly on IOU overlap. This is challenging here as the BumbleBox (and many other camera trap setups) uses a lower frame rate (4.5 fps) and a fast-moving bee may have little or none of its body overlap between consecutive frames, making IOU-based matching less reliable. In addition, many biological applications require long-term retention of individual identity, including when individual animals come in and out of view or are temporarily obscured, creating important challenges for long-term multi-object tracking.

\subsection{AI-assisted Annotation}
AI-assisted annotation tools commonly allow users to manually label an initial subset of data, train a preliminary model, and use its predictions to accelerate subsequent annotation through iterative refinement \cite{CVAT_ai_Corporation_Computer_Vision_Annotation_2023,LabelStudio,smith2022rootpainter,X-AnyLabeling}. However, our workflow required a combination of features not jointly available in existing tools: video navigation and frame extraction, SAM2-assisted mask generation \cite{ref:ravi2025sam2}, iterative retraining of detection and segmentation models during annotation, and free access. Therefore, developed a custom annotation GUI inspired by existing platforms (\cite{X-AnyLabeling,CVAT_ai_Corporation_Computer_Vision_Annotation_2023} in particular). It supports video-based inspection, SAM2-assisted coarse masks, manual mask correction, and iterative YOLO retraining.

\section{Dataset}
\label{sec:dataset}

\subsection{Training and Validation Dataset}
Frames from 224 Videos from 5 different microcolony experiments run across the past 7 years (including the one we analyze results on) were annotated and split into training and validation sets. Videos from many experiments were used to improve generalization. Training and validation splits were defined at the video level, such that all annotated frames from a given video were assigned to the same split. This is intended to limit overfitting, since individual bees may remain nearly stationary across frames within a video. For instance segmentation results training and validation frames from the same video were at least 5 frames apart, to try to prevent overfitting.

\Cref{tab:dataset} provides an overview of the annotations. Four object classes were annotated: bees, pollen balls, nest structures (e.g. brood and wax pots), and chambers. Annotation was performed using a multi-stage AI-assisted workflow described in supplementary  \cref{sec:supp:annotationprocess}. 

For results on frame-to-frame tracking, 13 of the above validation videos were partially annotated for a total of 122 frames and 2,438 bee instances. Not all instances were segmented, some frames simply used bounding boxes allowing for quicker annotation. Further details about training and validation splits across data source for object detection and tracking is provided in supplementary \cref{sec:supp:trainvalsource}.

\subsection{Pesticide Dataset Generation}
We apply the BeeWhere models on an experimental dataset of queenless bumble bee (\textit{Bombus impatiens}) microcolonies. Microcolonies (n = 24) were created using established protocols  \cite{EastonCalabria2023}, each consisting of eight workers and accompanying brood (larvae and pupae). Microcolonies were exposed to a range of concentrations (0-100 ppb) of a common neonicotinoid pesticide (imidacloprid) in nectar over multiple days. The experiment consisted of three paired rounds, each with one microcolony exposed to each of eight unique concentrations. Data from only the first round were used in model training and validation. Example analyses presented in this paper focus on a subset of colonies (n = 3 control [0 ppb imidacloprid in nectar] microcolonies and n = 3 microcolonies exposed to 10 ppb imidacloprid in nectar). 

Each microcolony was housed in a custom imaging chamber measuring 144 mm x 76 mm x 50 mm and had \textit{ad libitum} access to nectar and pollen. Microcolonies were maintained at 27°C and 50\% relative humidity and imaging was conducted under IR (940 nm) illumination with 10-s videos collected every 10 min using a Raspberry Pi High Quality Camera and a 12mm Arducam lens at full resolution, 4056x3040. ArUco tracking was conducted following \cite{ref:easton2025bumblebox}.

Microcolonies were created from mature, queenright commercial \textit{B. impatiens} (BioBest) source colonies.  The brood masses for the six microcolonies ranged from 3.37 to 3.87 g, and was standardized to larvae and early-stage pupae at the time of microcolony creation. Eight workers were randomly selected from the corresponding source colony for each microcolony. Workers were individually marked with unique ArUco fiducial tags printed on waterproof plastic paper and attached to the thorax using cyanoacrylate glue, following \cite{EastonCalabria2023}. In total, 48 workers were tagged across the six focal microcolonies. Microcolonies were monitored for several days (2.7-4.6 d after introduction of treatment conditions), with variation across colonies in recording time due to hardware failures in recording equipment.

\begin{table}[tb]
  \caption{Training and Validation dataset statistics. Note that some frames have all class annotations and others do not. Pollen Balls, Nest Structures and Chambers are annotated per video since they are mostly static over the course of the 10s video.}
  \label{tab:dataset}
  \centering
  \begin{tabular}{@{}lrrrrrrrrr@{}}
    \toprule
    Class & \multicolumn{2}{c}{No. Videos} & \multicolumn{2}{c}{No. Frames} & \multicolumn{2}{c}{No. Bbox} & \multicolumn{2}{c}{No. Seg}\\
    & Train & Val & Train & Val & Train & Val & Train & Val\\
    \midrule
    Bee & 151 & 73 &  321 & 162 & 5,237 & 3,206 & 4,252 & 692 \\
    Pollen Ball & 69 & 12 & - & - & 256 & 38 & 256 & 38\\
    Nest Structure & 89 & 28 & - & - & 290 & 115 & 286 & 115\\
    Chamber & 55 & 17 & - & - & 98 & 33 & 97 & 33\\
  \bottomrule
  \end{tabular}
\end{table}

\section{Methods}
\subsection{Object Detection and Instance Segmentation}
We used YOLO26 \cite{hidayatullah2026yolo26}, implemented by Ultralytics \cite{jocher2026ultralyticsyolo26unifiedrealtime}, for object detection and instance segmentation. YOLO is a widely used and well-documented family of models for bounding-box detection and mask prediction \cite{arkin2023survey}. Models were initialized from COCO-pretrained weights \cite{lin2014microsoft} and trained with standard augmentations, including CutMix, MixUp, flips, rotations, and random cropping. Training ran for up to 1000 epochs with early stopping after 100 epochs without improvement in validation mAP$_{0.5:0.95}$, using an initial learning rate of 0.001. Non-maximum suppression was applied during inference. Full training configurations, hyperparameters, loss terms, and final trained models are provided in the accompanying GitHub repository.

Annotations were produced through the multistage process described in detail in supplemental \cref{sec:supp:annotationprocess}, including iterative retraining of a bounding-box model (YOLO-Bbox-Bee), using SAM2 to provide initial segmentations, a coarse segmentation model trained on full images (YOLO-Seg-Bee), and a fine-grained segmentation model trained on crops around bounding boxes (YOLO-Bee-Fine). Because not all annotated frames contained labels for every class, and because some target masks were allowed to overlap, we trained separate class-specific models. This allowed each model to use data selected for the most relevant variation: chamber annotations covered different microcolony configurations, hive annotations captured visually diverse hive appearances, and bee annotations focused on dense or ambiguous bee clusters. Class-specific training reduced the need for exhaustive annotation of every class in every frame and allowed new classes, such as pollen balls, to be added without retraining the computationally expensive bee model.

\paragraph{Object Detection Metrics}
Mean Average Precision ($mAP_{IOU}$) averages class-wise Average Precision (AP), which summarizes the precision--recall curve at specified Intersection-over-Union (IoU) thresholds. Because each model was trained on a single class, we report AP per model and compute overall mAP afterward.

\subsection{Behavioral Metrics}
In order to test the utility of our models, we analyze the 4,468 videos taken during the dosing experiments described above. We exclude 707 videos taken before dosing. Since analyzing the videos is time consuming, much of the analysis below focuses on a single frame from each video. To explore the potential for BeeWhere to assess key behaviors, we quantified several key behavioral metrics, including proximity of each bee to its nearest neighbor based on the segmentation masks which reflect the degree of social clustering\cite{Crall2018}. Finally, we demonstrate the utility of long-term markerless (tag-free) tracking by monitoring population growth (i.e. emergence of new workers) within chambers over the course of the observation period and plotting the spatial occupancy of the bees.

\subsection{Object Tracking}
In this paper we compare results on three tracking methods. The simplest is what we call \textit{centroid} tracking, which does hungarian matching using the distance between centroids of the current and previous frame with a maximum threshold. When the threshold is reached the instance is considered a new identification. In our case we chose to use a threshold of 200 pixels, as this seemed to be approximately the frame-to-frame distance of a fast moving bee. 

The other two methods compared are SimpleIOU with a threshold of 0.5, which just matches based on the highest Intersection-over-Union (IoU) between frames\cite{bochinski2017high}, and Bytetrack\cite{zhang2022bytetrack} with a confidence threshold of 0.5, high IoU threshold of 0.6 and low IoU threshold of 0.3, and lost tracking after 10 frames. 

We explore augmenting object tracking with reidentifications using ArUco markers. For this a forward pass is performed. At each frame the tracker checks if each segmentation overlaps with an ArUco detection in the frame, if it does, and that tracking instance does not already have a ArUco id associated with it, that ArUco id is assigned to that tracking instance. This is also used to reassign later tracks to the same instance, assuming that instance was lost. Instances which already have another ArUco id associated with them are ignored.

\paragraph{Object Tracking Metrics} Multiple Object Tracking Accuracy (MOTA)\cite{bernardin2008evaluating} measures overall tracking performance by penalizing false positives, false negatives, and identity switches, while Identity F1 (IDF1)\cite{ristani2016performance} measures how consistently object identities are preserved over time. Higher MOTA and IDF1 values indicate better tracking performance, with fewer detection, association, and identity-preservation errors.

In the results we report preliminary results on both how ArUco tags can improve the Object Tracking, and how the instance segmentations can be used to prolong the reidentification of individuals.

\section{Results}

\begin{figure}[tb]
    \centering
    \includegraphics[width=0.95\textwidth]{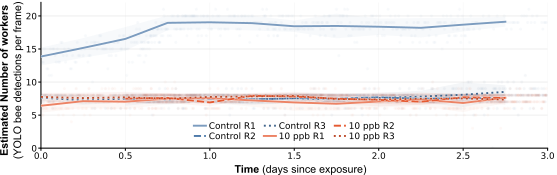}
    \caption{
      Number of YOLO detections per frame for each trial, shown with a rolling average. Control R1 showed an increase in population over the experiment.
    }
    \label{fig:yolo_detections_per_frame}
\end{figure}

\begin{maybefigure}[tbh!]
  \centering
  \begin{subfigure}{0.45\linewidth}
    \includegraphics[width=\textwidth]{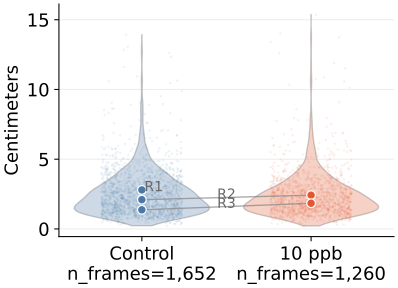}
    \caption{ArUco marker centroids}
    \label{fig:res:nearest:aruco}
  \end{subfigure}
  \hfill
  \begin{subfigure}{0.45\linewidth}
    \includegraphics[width=\textwidth]{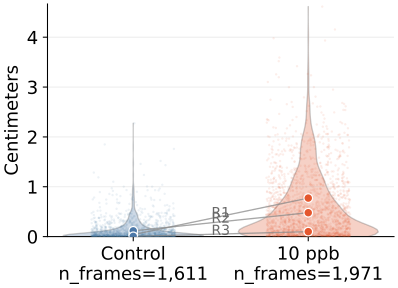}
    \caption{YOLO-Seg-Bee contours}
    \label{fig:res:nearest:inst}
  \end{subfigure}
  \caption{Distributions of estimated median distance to nearest bee per frame using (a) ArUco markers and (b) instance segmentations. n\_frames is the number of frames with detections included in the calculation. It should be noted that the R1 trial does not have any ArUco detections as the videos were too blurry to identify any, but bees were detected with the instance segmentation model. The mismatch in frame numbers between the two approaches for control is due to skipped frames which encountered errors during segmentation inference.}
  \label{fig:res:nearest}
\end{maybefigure}
\subsection{Object Detection and Instance Segmentation on Single Frames}
\begin{table}[tb]
  \centering
  \caption{Average precision (AP) results per model, and number of epochs run until early stopping. The results of the bounding box and fine grained bee model are shown for completeness but used only during annotation. The segmentation (`Seg') models are used for inference and further results are based on these models.}
  \label{tab:maps}
  \begin{tabular}{@{}llll@{}}
    \toprule
    Model & Val $AP_{0.5}$& Val $AP_{0.5:0.95}$ & No. Epochs\\
    \midrule
    YOLO-BBox-Bee & 0.9532 & 0.75067 & 323 \\ 
    YOLO-BBox-Bee + SAM2\footnotemark & 0.9343 & 0.4428 & --\\
    YOLO-BBox-Bee + YOLO-Bee-Fine & 0.9099 & 0.6466 & 490\\ 
    \hline
    YOLO-Seg-Bee & 0.97739 & 0.75688 & 240 \\ 
    YOLO-Seg-PollenBall & 0.93417 & 0.7432 & 221\\ 
    YOLO-Seg-Nest & 0.79717 & 0.55338 & 360\\ 
    YOLO-Seg-Chamber & 0.995 & 0.995 & 14 \\ 
    \hline
    mAP Across YOLO-Seg Models\footnotemark & 0.92593 & 0.76212 & --\\
    \bottomrule
  \end{tabular}
\end{table}
\footnotetext{Note this is the pretrained SAM2 without any fine tuning.}
\footnotetext{Here we calculate the mean average precision for the classes across the different `Seg' models. This is atypical but should provide some comparability of the general precision of the group of models.}

\Cref{tab:maps} reports on the accuracy of the object detection and instance segmentation. Most segmentation models achieved high performance ($mAP_{0.5:0.95}$ above 0.7), with exception of the nest segmentations. Lower performance on nest segmentations may occur for several reasons. First, the nest edges are often obscured in single frames by bees, and we completed the nest annotations at a video level (as the nest rarely measurably changes structure within a video, and we want to be able to predict when a bee is on top of a nest, so we want the model to guess what is underneath the bee). Second, we decided to treat disconnected nest components as their own instances, as we thought this might work better with YOLO's implementation of predicting extents of objects. Also the nest components are often confused with pollen balls, and these classes are often difficult for human observers to clearly distinguish, highlighting that this could be a fundamentally challenging vision task. Further testing would be needed to determine the true source of error and potential improve nest segmentation models. We suspect combining the nest structure and pollen ball models into one YOLO model could improve this, but we leave this for future work.

\subsubsection{Spatial Distribution of Detections}

\begin{maybefigure}[tb]
  \centering

    \centering
    \includegraphics[width=0.95\textwidth]{figures/bee_detections_spatial_maps.jpg}

    \caption{
      Spatial bee detections for each trial from ArUco and YOLO-Seg-Bee. Average chamber, nest structure, and pollen ball contours from 100 random frames are shown. Note that chambers moved slightly over time. This can bee seen in the 'Avg' column. Round 1's imidacloprid treatment is missing ArUco detections completely because the images were blurred. We can see from the 'Last' column that often nest structures and pollen balls are not entirely captured, partly due to pollen ball movement. 
    }
    \label{fig:locationmaps}

\end{maybefigure}


\Cref{fig:locationmaps} shows all detections across the 3,761 frames using ArUco tags only (columns 1 and 3) and instance segmentations (columns 2 and 4), for control (left) and 10ppb dosings (right). Overall, instance segmentation yielded ~2.5 times more bee detections (n = 34,423) across all trials than ArUco tracking (n = 12,618). This included one trial where no tags were tracked due to incidental camera blur (arising from experimental error), but where instance segmentation nonetheless showed strong performance (\Cref{fig:locationmaps}, Round 1, 10 ppb). Instance segmentation and tag-based tracking also show apparent differences in spatial distribution of detection, with instance segmentation showing higher detection rates of bees clustered on the center of the nest and on the brood structure, likely reflecting the higher rates of tag occlusion in the central, dense areas of the nest. This pattern is striking when comparing the tracking methods in the Control colony of Round 1. This microcolony had high rates of emergence of new workers during the trial, as seen in \cref{fig:yolo_detections_per_frame} , which could drive this pattern as newly emerged worker bees typically remain on the brood structure and focus on nursing for the first few days of their life. Because newly emerged bees are untagged and commonly remain near the brood, ArUco tracking alone does not capture many bees tending the nest. Instance segmentation detects these individuals, making their concentration around the nest structure apparent. \cref{fig:yolo_detections_per_frame} also demonstrates the potential for instance segmentation approaches to automatically monitor changes in colony size: Five of the six microcolonies remained near the initial count of eight tagged workers. In the remaining microcolony, the increase in BeeWhere detection counts was consistent with the observed emergence of new workers.



\subsubsection{Proximity to Other Bees}

\Cref{fig:res:nearest} shows a comparison of proximity between nearest bee calculated using ArUco tags (panel a) and using instance segmentation (panel b). Note the much higher range (y axis) of inter-bee distance based on tag tracking (panel a), reflecting higher average estimates from ArUco tags than markerless segmentation. This difference may arise from distinct effects. First, inter-bee distances based on segmentations will be necessarily lower than distances based on tags, as extremities (e.g. limbs) can come into closer physical contact than tags. Second, given that segmentation detects more bees, this will contribute to measurable reduction in estimated proximity to the nearest bee, particularly in high density, central areas of the nest as noted above. The two tracking approaches may also yield qualitatively different patterns. Using ArUco tags the distributions of distances are almost indistinguishable between experimental treatments, while segmentation detected a stronger difference between control and imidacloprid-exposed colonies. This could reflect improved detection of bees in close proximity to the nest center using segmentation, which in turn could be a strong driver of the difference between treatments. Further tests would need to be completed to determine the significance and generality of the pattern observed here in a limited number of replicate microcolonies. We also note that in the case of control round 1, since there were many newly emerged bees this median distance measure drops even further, so it may not prove to be an unbiased measure for microcolonies that exhibit population growth. Proximity to nest strucure results are shown in supplemental \cref{sec:supp:nest}.

\subsection{Object Tracking}
\begin{table}[tb]
  \caption{Tracking performance for segmentation and bounding box models measured using Multiple Object Tracking Accuracy (MOTA) and Identity F1 Score (IDF1). Best results are bolded.}
  \label{tab:tracking}
  \centering
  \begin{tabular}{@{}llccc@{}}
    \toprule
    Prediction Source & Tracking Method  & MOTA (\%) & IDF1 (\%) & ID Switches \\
    \midrule
    \multirow{3}{*}{Ground Truth} & ByteTrack & 89.7 & 89.7 & 322 \\
     & SimpleIOU & 93.7 & 93.7 & 196 \\
     & Centroid - 200 & \textbf{99.0} & \textbf{99.0} & \textbf{34} \\
    \midrule
    \multirow{3}{*}{YOLO-BBox-Bee} & ByteTrack & 84.7 & 88.3 & 233 \\
     & SimpleIOU & 87.9 & 91.5 & 138 \\
     & Centroid - 200  & 90.9 & 91.5 & 44 \\
    \multirow{3}{*}{YOLO-Seg-Bee} & ByteTrack & 79.1 & 82.6 & 422 \\
     & SimpleIOU & 83.5 & 87.1 & 281 \\
     & Centroid - 200 & 91.0 & 94.7 & 46 \\
    \multirow{3}{*}{YOLO-Seg-Bees + Aruco} & ByteTrack & 82.8 & 86.3 & 311 \\
     & SimpleIOU & 85.8 & 89.4 & 212 \\
     & Centroid - 200  & \textbf{91.2} & \textbf{94.9} & \textbf{40} \\
    
  \bottomrule
  \end{tabular}
\end{table}

\begin{table}[tb]
  \caption{Comparison of the number of bees detected using the Segmentation model only; the identified ArUco tags only; and their combination on the validation tracking dataset. FPs (False Positives) are wrong identifications due to ID switches. FNs (False Negatives) are identifications we failed to pick up, but could have with perfect tracking and identification. Results generated using Centroid - 200 tracking algorithm.}
  \label{tab:trackingindiv}
  \centering
  \begin{tabular}{@{}lcccc@{}}
    \toprule
    Method & Detections & ArUco Identified Detections & FPs & FNs\\
    \midrule
    ArUco & 915 &  915 & - & - \\
    Yolo-Seg-Bee & 2378 & 0 & - & -  \\
    ArUco + Yolo-Seg-Bee & 2378 & 1110 & 4 & 32 \\
    Ground truth & 2441 & 1142 &  - & - \\
  \bottomrule
  \end{tabular}
\end{table}

\Cref{tab:tracking} shows a comparison of different tracking algorithms on the validation set. Overall the combination of segmentation and Aruco using centroid matching with a max distance of 200 pixels performed the best. And tests on ground truth show that the performance gap to perfect is largely due to object detection performance. Although the increase in number of ID switches indicates some room for improvement on the tracking front as well.

\Cref{tab:trackingindiv} shows that instance segmentation has the potential to double the number of detections compared to ArUco detection alone. And while ArUco combined with instance segmentations allows us to increase the number of identified (known) detections from 915 to 1110, we also find false positive detections (detections assigned the wrong ArUco tag due to imperfections in the segmentation and tracking process). So use of this method to extend re-identification is dependent on the level of accuracy needed. Though the number of false positives found is small (4) because the validation is preliminary and  was only run on a maximum of 11 consecutive frames, if the tracking were used on longer segments, we would expect to see more false positives. Further testing would be required to determine exactly how beneficial segmentations are to extending ArUco identifications across different contexts and experimental designs.

\section{Conclusion}
In this paper, we demonstrate how instance segmentations for BumbleBox images can be generated through a AI-assisted annotation workflow. We provide the accompanying software and trained segmentation model to support reproducible annotation and analysis. We find that this approach can substantially improve overall detection and tracking of bees, and allows for automated behavioral inference at both the individual- and colony-scales. Instance segmentation substantially boosted overall detection rates in the validatation dataset and improved performance on a multi-object tracking task when combined with tag-based identification. Overall, the BeeWhere framework has enormous potential to provide automated detection and tracking of uniquely identified bees within nests, as well as automated segmentation of nest elements (e.g. food resources and nest structure) that are critical for behavioral inference. BeeWhere expands the existing toolkit of tracking approaches by providing robust detection and segmentation even when bees and tags are partially obscured under dense nest conditions. Markerless segmentation of bees and nest elements can also provide scalable, non-invasive quantification of structural changes within the nest (e.g. brood construction) and population growth (e.g. counts of newly emerged workers) that are especially important for linking behavior to colony -scale outcomes (e.g. growth and reproduction). 

This work also demonstrates the potential of this approach by quantifying fine-grained behavioral metrics to assess the effects of pesticide exposure on bumblebee behavior. Specifically, we analyzed inter-bee distance and bee detection as proxies for social interactions and population growth. Our results indicate exposure effects on social behavior, as expected from current literature, but our method allows further granularity, allowing us to see distributional differences in between-bee distances. These findings illustrate the potential of instance-level image analysis for detecting subtle behavioral effects in controlled colony-monitoring experiments. However, we emphasize that the present study is exploratory. Because only three microcolonies of eight workers were analyzed per dose condition, the results should not be interpreted as conclusive evidence of dose-specific pesticide effects. Additional experiments are required to assess repeatability, separate treatment effects from colony-specific variation, and support stronger statistical conclusions.

\section{Ethical Concerns}
This work raises ethical considerations related to surveillance, animal welfare, large language models, and computational cost. Although our methods are intended for insect behavioral analysis, animal-tracking tools can in principle be adapted for human surveillance; we discourage such use and emphasize that our contribution applies already widespread tracking technologies to pollinator research. The experiments involve rearing, monitoring, and pesticide exposure of bees, raising animal-welfare concerns, but aim to support research that may ultimately reduce pesticide-related harm to pollinators and ecosystems. Large language models, including Claude Sonnet, DeepSeek, and Codex, assisted with GUI development, analysis code, and manuscript editing, and we acknowledge unresolved concerns around training data and copyright. Finally, model training and analysis have a carbon footprint: Carbontracker~\cite{anthony2020carbontracker} was run for a single epoch of each model, yielding an estimated 540g CO2eq for full training of the final models; the footprint of full experimental tracking and human-in-the-loop annotation was not measured.
\bibliographystyle{splncs04}
\bibliography{main}




\clearpage
\appendix

\section{AI-Assisted Annotation Process}
\label{sec:supp:annotationprocess}
The multistage AI-Assisted annotation process is described here. First, objects were manually annotated with bounding boxes, which were used both to train an initial bee detector and to prompt SAM2\cite{ref:ravi2025sam2} for coarse mask generation. The resulting masks were manually refined to correct boundary errors and missed regions. For bee instances, we additionally trained a crop-based fine-grained YOLO26 segmentation model (YOLO-Bee-Fine) on cropped images of bees (maintaining full resolution), which provided higher-quality masks than were feasible when training directly on the original 12-megapixel frames using the available local hardware. Although YOLO was not designed specifically for this single-instance crop-based setting, it allowed us to reuse the existing pipeline and produced masks of sufficient quality for annotation refinement. For inference on complete images, we trained a coarse-grained YOLO26 instance segmentation model on full frames at half resolution on a larger GPU. An overview of the workflow and example segmentation quality are shown in \cref{fig:annotation_pipeline}.

We did not record annotation times and therefore do not benchmark the efficiency of this workflow against other AI-assisted annotation methods. Instead, we report the procedure to document dataset construction and support reuse in related BumbleBox studies.

The final training and validation dataset contains 483 frames from five microcolony experiments conducted over the past 7 years; dataset statistics are reported in \cref{tab:dataset} and \cref{tab:supp:datasources}. Although evaluation is performed on a single standardized setup, training data were drawn from multiple experiments to improve generalization across BumbleBox configurations. All training data are released with this paper.

\begin{maybefigure}[t]
\centering
\resizebox{\textwidth}{!}{%
\begin{tikzpicture}[
  font=\figfont,
  stage/.style={
    rounded corners=3mm,
    draw=#1,
    line width=0.8pt,
    fill=#1!5,
    minimum height=9.8cm,
    inner sep=2.5mm
  },
  subbox/.style={
    rounded corners=2mm,
    draw=#1,
    line width=0.65pt,
    fill=#1!4,
    inner sep=1.8mm
  },
  arrow/.style={
    -{Latex[length=2.2mm]},
    line width=0.8pt,
    draw=black!70
  },
  smallarrow/.style={
    -{Latex[length=1.8mm]},
    line width=0.65pt,
    draw=black!70
  },
  model/.style={
    rounded corners=1.5mm,
    draw=#1,
    fill=#1!5,
    align=center,
    minimum width=1.25cm,
    minimum height=0.65cm
  },
  thumb/.style={
    draw=black!35,
    rounded corners=1mm,
    fill=black!3,
    minimum width=2.5cm,
    minimum height=1.35cm
  }
]


\node[stage=stageblue, minimum width=3.05cm] (s1) at (0,0) {};
\node[stage=stageblue, minimum width=3.05cm, right=0.55cm of s1] (s2) {};
\node[stage=stageblue, minimum width=3.05cm, right=0.55cm of s2] (s3) {};
\node[stage=stageblue, minimum width=3.05cm, right=0.55cm of s3] (s4) {};

\node[anchor=north west, font=\bfseries\small, align=left]
  at ($(s1.north west)+(0.18,-0.18)$) {\textcolor{stageblue}{\large 1.} BBox\\Annotations};

\node[anchor=north, font=\bfseries\figtiny] at ($(s1.north)+(0,-1.1)$) {Input frame};
\node[thumb, anchor=north] (img1) at ($(s1.north)+(0,-1.4)$) {%
  \includegraphics[width=2.35cm,height=1.25cm,keepaspectratio]{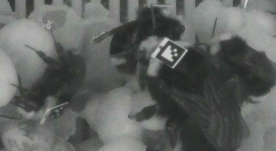}%
};

\node[anchor=north, font=\bfseries\figtiny] at ($(img1.south)+(0,-0.45)$) {Manual Annotations};
\node[thumb, anchor=north] (img1b) at ($(img1.south)+(0,-0.75)$) {%
  \includegraphics[width=2.35cm,height=1.25cm,keepaspectratio]{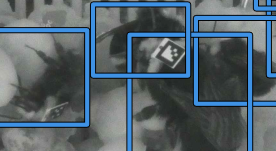}%
};
\draw[smallarrow] ($(img1.south)+(0,-0.02)$) -- ($(img1b.north)+(0,0.2)$);

\node[anchor=north, align=center] (s1icons) at ($(img1.south)+(0.13,0.00)$)
{\BeeIcon[0.2]\enskip\PollenIcon[0.4]\quad\HiveIcon[0.3]\enskip\ChamberIcon[0.5]\\[-0.2ex]
   };

\node[anchor=north, font=\bfseries\figtiny] at ($(img1b.south)+(0,-0.45)$) {Train YOLO Model};
\node[thumb, anchor=north] (img1c) at ($(img1b.south)+(0,-0.75)$) {%
  \includegraphics[width=2.35cm,height=1.25cm,keepaspectratio]{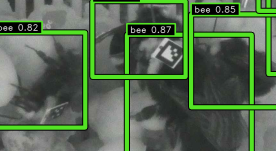}%
};
\draw[smallarrow] ($(img1b.south)+(0,-0.02)$) -- ($(img1c.north)+(0,0.35)$);
\draw[smallarrow] ($(img1c.north)+(0.15,0.35)$) -- ($(img1b.south)+(0.15,-0.02)$);

\node[anchor=north, align=center] (s1icons) at ($(img1b.south)+(-0.35,0.00)$)
{\BeeIcon[0.2]\\[-0.2ex]
   };

\node[align=center, anchor=south] at ($(s1.south)+(0,0.35)$)
  {Manually-drawn boxes \\ for bees, pollen balls, \\ hives, and chambers.\\ Used for bees to \\ train YOLO model and \\ annotate faster and \\retrain YOLO, etc.};

\node[anchor=north west, font=\bfseries\small, align=left]
  at ($(s2.north west)+(0.18,-0.18)$) {\textcolor{stageblue}{\large 2.} SAM2 \\Segmentations};

\node[anchor=north, font=\bfseries\figtiny] at ($(s2.north)+(0,-1.1)$) {BBoxes};
\node[thumb, anchor=north] (img2) at ($(s2.north)+(0,-1.4)$) {%
  \includegraphics[width=2.35cm,height=1.25cm,keepaspectratio]{images/annotation_process/03_yolo_bbox.png}%
};
\node[anchor=north, align=center] (s2icons) at ($(img2.south)+(0.13,0.00)$)
{\BeeIcon[0.2]\enskip\PollenIcon[0.4]\quad\HiveIcon[0.3]\enskip\ChamberIcon[0.5]\\[-0.2ex]
   };

\node[anchor=north, font=\bfseries\figtiny] at ($(img2.south)+(0,-0.45)$) {SAM2 Segmentations};
\node[thumb, anchor=north] (img2b) at ($(img2.south)+(0,-0.75)$) {%
  \includegraphics[width=2.35cm,height=1.25cm,keepaspectratio]{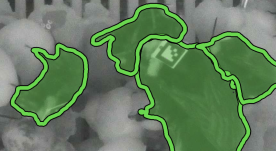}%
};
\draw[smallarrow] ($(img2.south)+(0,-0.02)$) -- ($(img2b.north)+(0,0.25)$);
\node[anchor=north, align=center] (s2icons) at ($(img2b.south)+(0.13,0.00)$)
{\BeeIcon[0.2]\enskip\PollenIcon[0.4]\quad\HiveIcon[0.3]\enskip\ChamberIcon[0.5]\\[-0.2ex]
   };

\node[anchor=north, font=\bfseries\figtiny] at ($(img2b.south)+(0,-0.45)$) {Manual Fixes};
\node[thumb, anchor=north] (img2c) at ($(img2b.south)+(0,-0.75)$) {%
  \includegraphics[width=2.35cm,height=1.25cm,keepaspectratio]{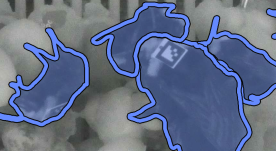}%
};
\draw[smallarrow] ($(img2b.south)+(0,-0.02)$) -- ($(img2c.north)+(0,0.25)$);

\node[align=center, anchor=south] at ($(s2.south)+(0,0.35)$)
  {YOLO bounding boxes \\ inform SAM2 masks \\ which are manually\\corrected.};

\node[anchor=north west, font=\bfseries\small, align=left]
  at ($(s3.north west)+(0.18,-0.18)$) {\textcolor{stagegreen}{\large 3.} Fine Bee\\Model};

\node[anchor=north, font=\bfseries\figtiny] at ($(s3.north)+(0,-1.1)$) {Manual Fixes};
\node[thumb, anchor=north] (img3) at ($(s3.north)+(0,-1.4)$) {%
  \includegraphics[width=2.35cm,height=1.25cm,keepaspectratio]{images/annotation_process/07_ground_truth.png}%
};
\node[anchor=north, align=center] (s1icons) at ($(img3.south)+(-0.35,0.00)$)
{\BeeIcon[0.2]\\[-0.2ex]
   };

\node[anchor=north, font=\bfseries\figtiny] at ($(img3.south)+(0,-0.45)$) {Train YOLO-Bee-Fine};
\node[thumb, anchor=north] (img3b) at ($(img3.south)+(0,-0.75)$) {%
  \includegraphics[width=2.35cm,height=1.25cm,keepaspectratio]{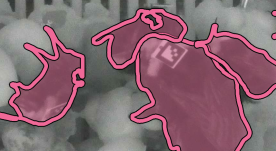}%
};
\draw[smallarrow] ($(img3.south)+(0,-0.02)$) -- ($(img3b.north)+(0,0.25)$);
\draw[smallarrow] ($(img3b.north)+(0.15,0.35)$) -- ($(img3.south)+(0.15,-0.02)$);

\node[align=center, anchor=south] at ($(s3.south)+(0,0.35)$)
  {A fine-grained YOLO\\model (YOLO-Bee-Fine)\\is trained on zoomed-in\\crops around the bee\\segmentations.\\This model is\\used for fine\\annotations and better\\at picking up\\small structures\\than YOLO-Seg-Bee,\\but can confuse close\\instances, so is not\\good for inference.};

\node[anchor=north west, font=\bfseries\small, align=left]
  at ($(s4.north west)+(0.18,-0.18)$) {\textcolor{stagegreen}{\large 4.} Coarse\\Model};

\node[anchor=north, font=\bfseries\figtiny] at ($(s4.north)+(0,-1.1)$) {Manual Fixes};
\node[thumb, anchor=north] (img4) at ($(s4.north)+(0,-1.4)$) {%
  \includegraphics[width=2.35cm,height=1.25cm,keepaspectratio]{images/annotation_process/07_ground_truth.png}%
};
\node[anchor=north, align=center] (s4icons) at ($(img4.south)+(0,0)$)
{\BeeIcon[0.2]\enskip\PollenIcon[0.4]\qquad\HiveIcon[0.3]\enskip\ChamberIcon[0.5]\\[-0.2ex]
   };

\node[anchor=north, font=\bfseries\figtiny] at ($(img4.south)+(0,-0.45)$) {Train YOLO-Seg-Bee};
\node[thumb, anchor=north] (img4b) at ($(img4.south)+(0,-0.75)$) {%
  \includegraphics[width=2.35cm,height=1.25cm,keepaspectratio]{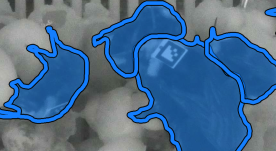}%
};
\draw[smallarrow] ($(img4.south)+(-0.15,-0.02)$) -- ($(img4b.north)+(-0.15,0.25)$);
\draw[smallarrow] ($(img4b.north)+(0.0,0.35)$) -- ($(img4.south)+(0.0,-0.02)$);

\node[align=center, anchor=south] at ($(s4.south)+(0,0.35)$)
  {A coarse-grained YOLO\\model (YOLO-Seg-Bee)\\is trained on\\the full images, one\\model per class.\\This model is\\used for inference\\on full frames.};

\draw[arrow] (s1.east) -- (s2.west);
\draw[arrow] (s2.east) -- (s3.west);
\draw[arrow] (s3.east) -- (s4.west);


\node[
  rounded corners=2mm,
  draw=black!25,
  fill=lightgraybg,
  minimum width=11cm,
  minimum height=0.72cm,
  anchor=north
] (legend) at ($(s2.south east)+(0,-0.45)$) {};

\node[anchor=west] at ($(legend.west)+(0.35,0)$)
  {\BeeIcon[0.3]\quad Bees};
\node[anchor=west] at ($(legend.west)+(2.65,0)$)
  {\PollenIcon[0.7]\quad Pollen Balls};
Input\node[anchor=west] at ($(legend.west)+(5.45,0)$)
  {\HiveIcon[0.5]\quad Hives};
\node[anchor=west] at ($(legend.west)+(7.65,0)$)
  {\ChamberIcon[0.9]\quad Chambers};

\end{tikzpicture}%
}
\caption{
Overview of the staged annotation pipeline. Arrows going both ways represent an iterative process of train-predict-correct-repeat. Icons next to each step represent which classes were included in the process. Since the bees are the hardest to predict and also have small structures (antennae and legs) which are important for social behavior, and therefore the most time was spent annotating them, therefore they have extra steps (training Yolo-Bee-Fine) to reduce the human effort involved in the annotation process. 
}
\label{fig:annotation_pipeline}
\end{maybefigure}

\clearpage
\section{Training and Validation Data Sources}
\label{sec:supp:trainvalsource}
Training and validation data was taken from 5 different microcolony experiments across the last 7 years, \cref{tab:supp:datasources} gives an overview of how training and validation were distributed for object detection, and \cref{tab:supp:datasourcestracking} gives an overview of how they were distributed for tracking. \Cref{fig:supp:datasources} shows an example frame from each datasource. The validation data is dominated by E2026, which is intentional since this is the dosing dataset we present results on, so we wanted to be certain that we have an idea how well it performs on that dataset in particular, especially since one of the rounds of data was blurry. 

\begin{table}[h!]
  \caption{Object Detection: Data sources for training and validation data. Each row is a different experiment run on a bumblebee microcolony. Note that the extents of the segmentations also count as bounding boxes, hence the number of bounding boxes always includes the segmentations. }
  \label{tab:supp:datasources}
  \centering
  \begin{tabular}{@{}lccccccc@{}}
    \toprule
     & & \multicolumn{2}{c}{No. Videos} & \multicolumn{4}{c}{No. Frames} \\
     & & & & \multicolumn{2}{c}{Bounding Boxes} & \multicolumn{2}{c}{Segmentations} \\
    Data Source & Year & Train & Val & Train & Val & Train & Val \\
    \midrule
    A2019 & 2019 & 33 & 4 & 54 & 4 & 54 & 4 \\
    B2021 & 2021 & 42 & 7 & 57 & 10 & 36 & 4 \\
    C2022 & 2022 & 12 & 5 & 32 & 4 & 24 & 1 \\
    D2024 & 2024 & 37 & 9 & 111 & 14 & 108 & 5 \\
    E2026 & 2026 & 27 & 48 & 67 & 130 & 62 & 23 \\
    \midrule
    Total & & 151 & 73 & 321 & 162 & 284 & 37 \\
  \bottomrule
  \end{tabular}
\end{table}

\begin{table}[h!]
  \caption{Object Tracking: Data sources for training and validation data. }
  \label{tab:supp:datasourcestracking}
  \centering
  \begin{tabular}{@{}lcccc@{}}
    \toprule
     & & No. Videos & \multicolumn{2}{c}{No. Frames} \\
     & & & Bounding Boxes & Segmentations \\
    Data Source & Year & Val  & Val  & Val \\
    \midrule
    A2019 & 2019 & 2 & 22 & 22 \\
    B2021 & 2021 & 1 & 10 & 0  \\
    C2022 & 2022 & 0 & 0 & 0  \\
    D2024 & 2024 & 1 & 11 & 11 \\
    E2026 & 2026 & 9 & 79 & 13 \\
    \midrule
    Total &  & 13 & 122 & 46 \\
  \bottomrule
  \end{tabular}
\end{table}

\begin{maybefigure}[h!]
  \centering
  \begin{subfigure}{0.45\linewidth}
    \includegraphics[width=\textwidth]{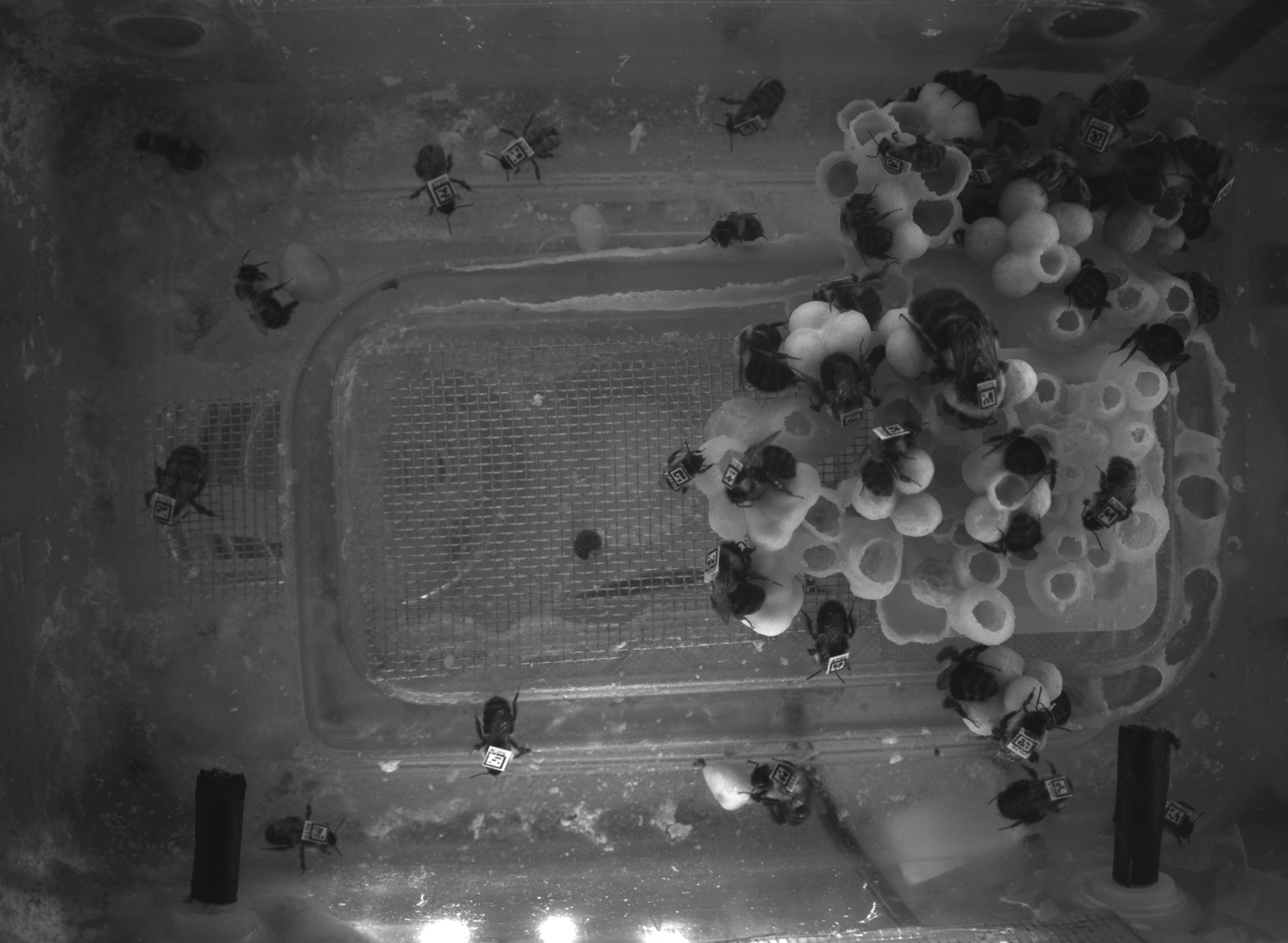}
    \caption{A2019}
  \end{subfigure}
  \hfill
  \begin{subfigure}{0.45\linewidth}
    \includegraphics[width=\textwidth]{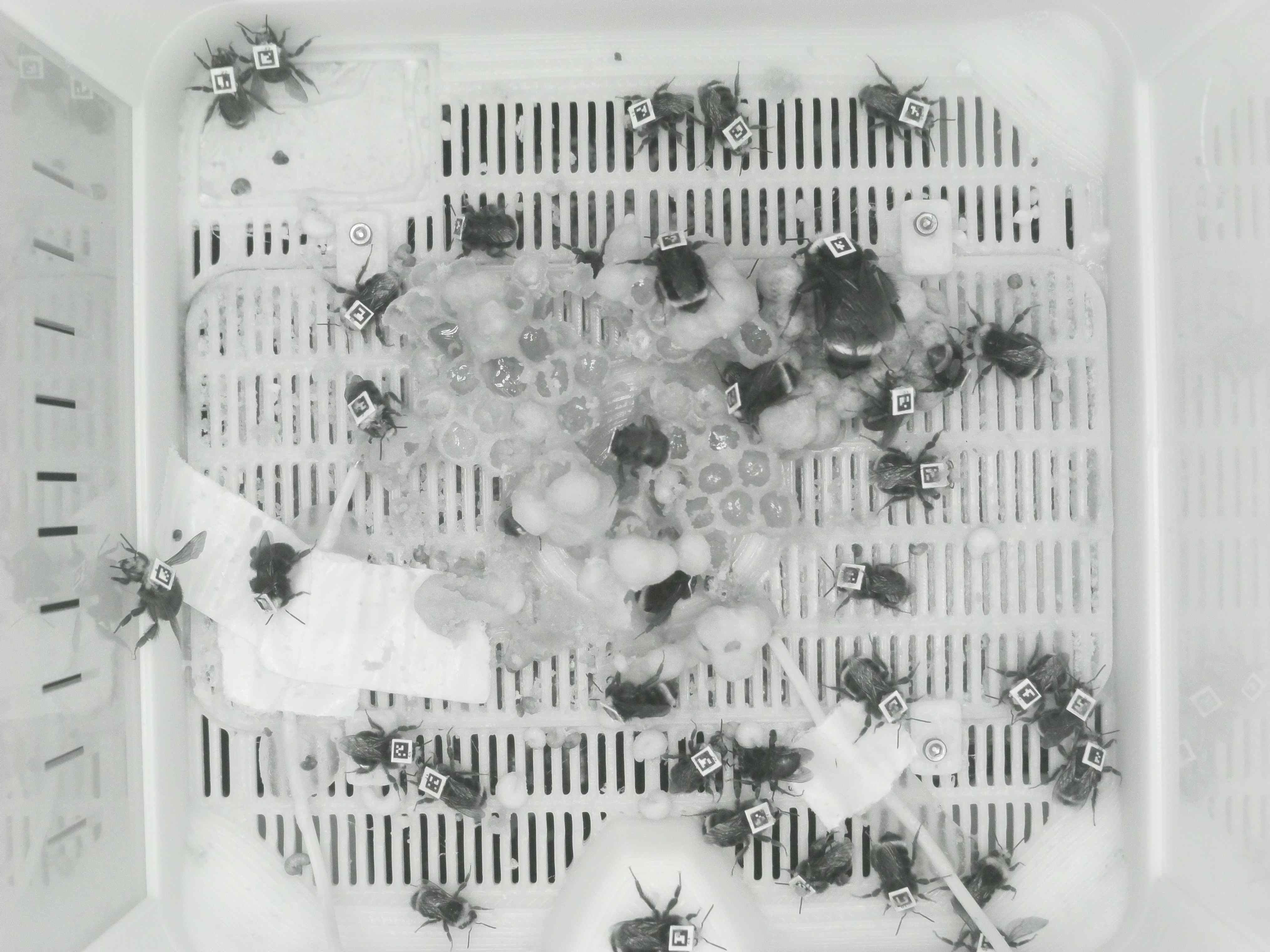}
    \caption{B2021}
  \end{subfigure}
  \\
  \begin{subfigure}{0.45\linewidth}
    \includegraphics[width=\textwidth]{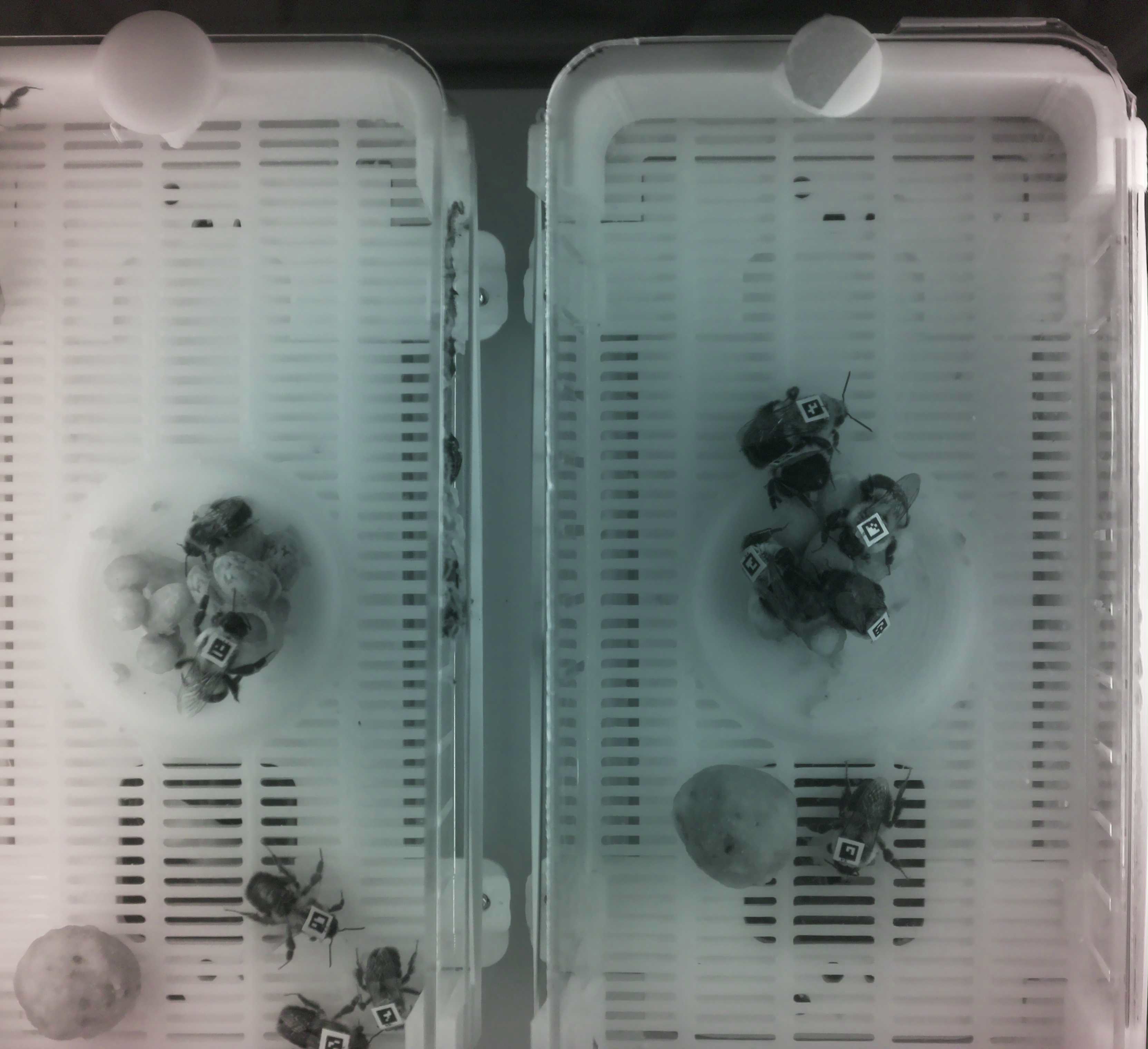}
    \caption{C2022}
  \end{subfigure}
  \hfill
  \begin{subfigure}{0.45\linewidth}
    \includegraphics[width=\textwidth]{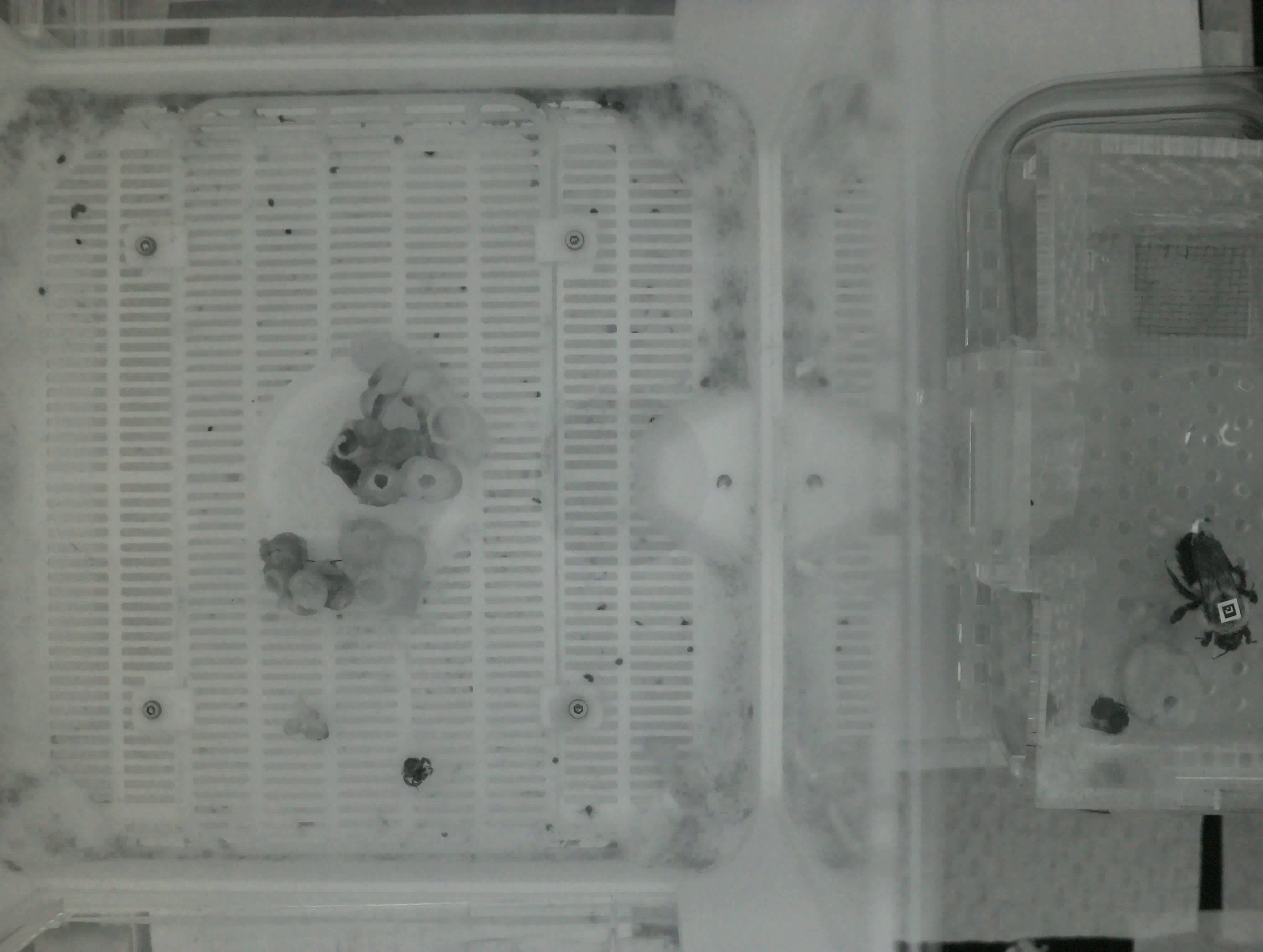}
    \caption{D2024}
  \end{subfigure}
  \\
  \begin{subfigure}{0.45\linewidth}
    \includegraphics[width=\textwidth]{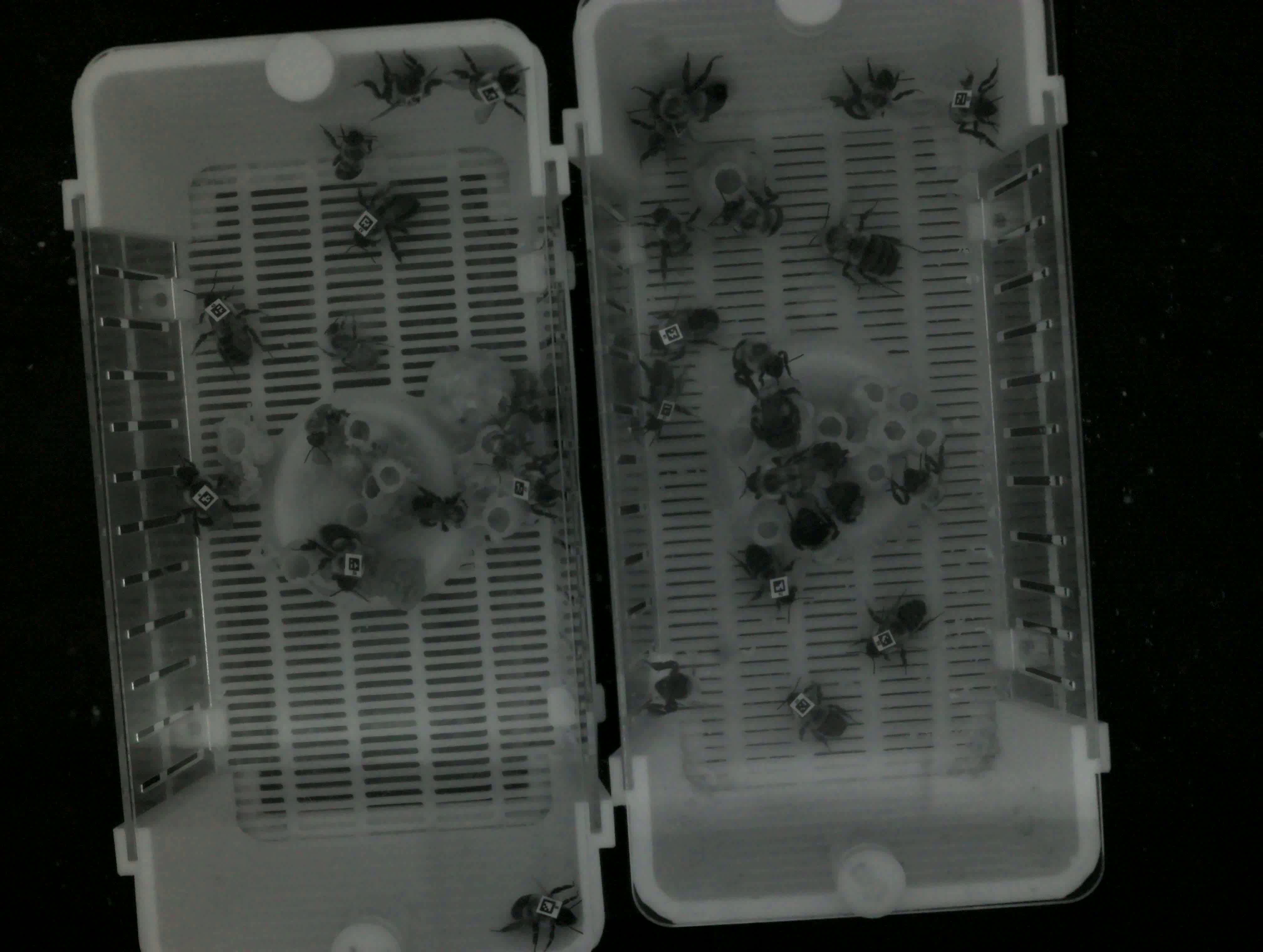}
    \caption{E2026}
  \end{subfigure}
  \hfill
  \\
  
  \caption{Example frames from training and validation data sources showing a variety of chamber and lighting arrangements.}
  \label{fig:supp:datasources}
\end{maybefigure}

\clearpage
\section{Proximity to Nest Structure}
\label{sec:supp:nest}
\begin{maybefigure}[h!]
  \centering
    \includegraphics[width=0.6\textwidth]{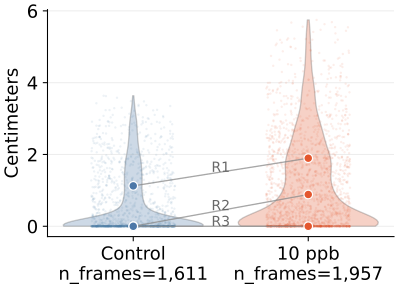}
  \caption{Distributions of estimated distance to the nest across treatments, predicted using segmentations}
  \label{fig:res:nest}
\end{maybefigure}

\Cref{fig:res:nest} shows a slight difference in estimated distribution of distance to nearest nest component, with 10ppb trials showing a greater tendency to be further from the nest. This pattern is consistent with previous work showing that exposure to imidacloprid increases distance from the nest center and reduces rates of nursing and brood care in bumble bee colonies \cite{Crall2018}. It should be noted that the nest prediction model was the worst performing model (see \cref{tab:maps}), so improvement to the model could potentially increase the

\end{document}